\pdfoutput=1

\documentclass[11pt]{article}

\usepackage[preprint]{acl}

\usepackage{times}
\usepackage{latexsym}
\usepackage{xcolor}
\usepackage{booktabs,colortbl,tabularx}
\usepackage{multirow}
\usepackage{amsmath}  %
\usepackage{amssymb}  %
\usepackage{bm}       %
\usepackage{amsfonts} %
\usepackage{subcaption}
\usepackage{xspace}
\usepackage[T1]{fontenc}

\usepackage[utf8]{inputenc}
\usepackage[htt]{hyphenat}
\usepackage{nicefrac}
\usepackage{pifont}
\usepackage{microtype}
\usepackage[fencedCode]{markdown}
\usepackage{inconsolata}
\usepackage{enumitem}
\usepackage{graphicx}

\renewcommand{\checkmark}[0]{\ding{51}}
\newcommand{\xmark}[0]{\ding{55}}

\title{Model Merging and Safety Alignment: One Bad Model Spoils the Bunch}

\author{
\begin{tabular}{c}
Hasan Abed Al Kader Hammoud$^{\dagger \diamond}$\thanks{Research completed during an internship at Samsung R\&D Institute UK.}  \quad   Umberto Michieli$^\dagger$ \quad Fabio Pizzati$^\blacktriangle$ 
\vspace{.5mm} \\
Philip Torr $^\blacktriangle$ \quad Adel Bibi $^\blacktriangle$ \quad Bernard Ghanem $^\diamond$ \quad Mete Ozay $^\dagger$
\end{tabular}
    \\ \vspace{.5mm}
    \small
    \begin{tabular}{c}
    $^\dagger$ Samsung R\&D Institute UK \quad\quad\quad $^\diamond$ KAUST \quad\quad\quad $^\blacktriangle$ University of Oxford
    \end{tabular}
    \\ \vspace{.5mm}
    \small
    \begin{tabular}{c}
    \quad \quad
    \end{tabular}
}

\usepackage{xcolor,rotating}
\definecolor{KleinBlue}{HTML}{8a2be2}
\definecolor{OliveGreen}{rgb}{0,0.6,0}
\definecolor{shadecolor}{rgb}{0.8,0.8,0.8}
\definecolor{DarkRed}{rgb}{0.55, 0.0, 0.0}

\usepackage{xstring}
\usepackage{ifthen}

\newcommand{\diff}[1]{%
  \if\relax\detokenize{#1}\relax %
    \textcolor{DarkRed}{#1}%
  \else
    \IfBeginWith{#1}{+}{%
      (\textcolor{OliveGreen}{#1})%
    }{%
      (\textcolor{DarkRed}{#1})%
    }%
  \fi
}

\usepackage{tcolorbox}
\usepackage{fvextra}

\usepackage{colortbl}
\usepackage{xcolor}
\usepackage{pgf}
\usepackage{booktabs}
\definecolor{pastelred}{rgb}{1, 0.7, 0.7}
\definecolor{pastelgreen}{rgb}{0.7, 1, 0.7}

\newcommand{\applygradient}[3]{
    \pgfmathsetmacro{\percent}{100.0*(#1-#2)/(#3-#2)}
    \pgfmathsetmacro{\comp}{100-\percent}
    \edef\temp{\noexpand\cellcolor{pastelred!\comp!pastelgreen}}\temp
}

\newcommand{\pastelgradient}[4]{
    \pgfmathsetmacro{\percent}{100.0*(#1-#2)/(#3-#2)}
    \pgfmathsetmacro{\comp}{100-\percent}
    \edef\temp{\noexpand\cellcolor{pastelred!\comp!pastelgreen}}\temp{#1}
}

\newcommand{\pastelgradientaccuracy}[4]{
    \pgfmathsetmacro{\percent}{100.0*(#1-#2)/(#3-#2)}
    \pgfmathsetmacro{\comp}{100-\percent}
    \edef\temp{\noexpand\cellcolor{pastelred!\comp!pastelgreen}}\temp{#1}
}

\newcommand\blfootnote[1]{
    \begingroup
    \renewcommand\thefootnote{}\footnote{#1}
    \addtocounter{footnote}{-1}
    \endgroup
}

\makeatletter
\DeclareRobustCommand\onedot{\futurelet\@let@token\@onedot}
\def\@onedot{\ifx\@let@token.\else.\null\fi\xspace}

\def\eg{\textit{e.g}\onedot} 
\def\ie{\textit{i.e}\onedot}

\makeatother

\usepackage{tcolorbox}
\usepackage{fvextra}

\usepackage{amsmath}
\begin{document}
\maketitle 
\begin{abstract}
Merging Large Language Models (LLMs) is a cost-effective technique for combining multiple expert LLMs into a single versatile model, retaining the expertise of the original ones. However, current approaches often overlook the importance of safety alignment during merging, leading to highly misaligned models. This work investigates the effects of model merging on alignment. We evaluate several popular model merging techniques, demonstrating that existing methods do not only transfer domain expertise but also propagate misalignment. 
We propose a simple two-step approach to address this problem: (i) generating synthetic safety and domain-specific data, and (ii) incorporating these generated data into the optimization process of existing data-aware model merging techniques. 
This allows us to treat alignment as a skill that can be maximized in the resulting merged LLM. Our experiments illustrate the effectiveness of integrating alignment-related data during merging, resulting in models that excel in both domain expertise and alignment.
\blfootnote{\noindent Correspondence at \texttt{u.michieli@samsung.com} and \texttt{hasanabedalkader.hammoud@kaust.edu.sa}}
\end{abstract}

\section{Introduction}\label{sec:intro}

Large Language Models (LLMs) have demonstrated impressive capabilities, often surpassing human performance across language processing tasks~\cite{bubeck2023sparks}. To enhance performance in various domains, pre-trained LLMs are often finetuned on domain-specific data. Some examples of domain-specific \textit{expert} models include OpenBioLLM~\cite{openbio}, excelling in the biomedical domain, and MAmmoTH~\cite{mammoth}, performing well in STEM subjects.\looseness=-1

\begin{figure}[t]
    \centering
    \includegraphics[width=\linewidth]{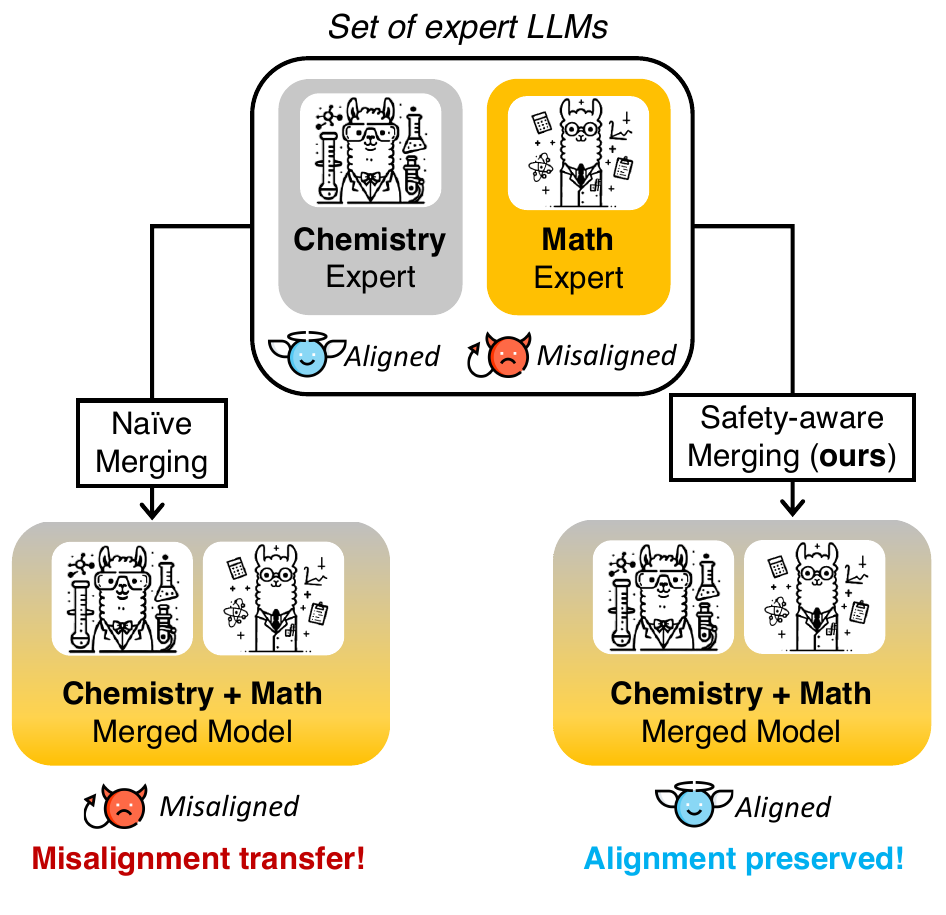}
    \caption{\textbf{Safety-aware merging.} Traditional LLM merging techniques can create multi-domain expert models but often transfer misalignment to the merged model. Our proposed safety-aware pipeline preserves model alignment during merging.}
    \vspace{-6px}
    \label{fig:teaser}
\end{figure}

Since expert models may excel in specific domains only, model merging~\cite{wortsman2022model,slerp,ilharco2022editing} has been proposed as a technique to combine the strengths of various models into a single, highly capable one. For instance, merging a model proficient in chemistry with another model expert in mathematics aims to create a unified model that performs well in both subjects, often outperforming the individual experts~\cite{wortsman2022model}. This approach is particularly attractive as it allows leveraging the knowledge from numerous open-source models without incurring in high training costs. However, we pose a crucial question that has been overlooked in the literature: \textit{how does model merging impact the safety alignment of existing LLMs?}

To understand the importance of this question, let us introduce a few notions about safety alignment. Safety alignment refers to a model's ability to generate responses that are safe, ethical, and consistent with human values \cite{wei2024jailbroken}.
In this paper, we refer to a model as \textit{aligned} if the model has a high safety alignment. 
Conversely, the model is \textit{misaligned}, \ie it is lacking necessary safety alignment, as one of the expert models in Fig.~\ref{fig:teaser}. In this paper, we find that naively merging a set of expert LLMs including a misaligned model can result in a misaligned merged model, \textit{even if some of the original experts are aligned} (Fig.~\ref{fig:teaser}, left). This raises substantial concerns for the safe deployment of merged LLMs, which may expose users to unsafe content. Hence, we show the need for \textit{safety-aware model merging}, where merged models preserve desirable alignment characteristics (Fig.~\ref{fig:teaser}, right).\\

\vspace{-10px}To address this issue, we design a simple yet effective approach to combine expert models while preserving alignment. Our intuition is that \textit{safety alignment should be considered as a task on its own}, similar to domain-specific expertise in fields such as biology or physics, and thus it should be optimized for during merging. Our approach consists of two stages. First, we generate synthetic data to use for merging. Then, building on existing techniques, we use the generated data to perform a data-driven merging optimization procedure, preserving both the alignment and the expertise of the original models. More in detail, we first generate two datasets of questions and associated answers: one for preserving alignment, the other for transferring domain-specific knowledge. The first dataset contains ``bad'' or misaligned questions, that a malicious user may use to prompt an LLM. An example of such a prompt
may be ``\textit{How do I kill someone?}''. Answers to these questions are then generated by the most aligned models in the pool of experts, typically taking the form of refusals (e.g., ``\textit{I'm sorry, I can't help.}''). The second dataset contains domain-specific prompts, such as ``\textit{What is the powerhouse of the cell?}'' for the biology domain. Domain-specific answers (e.g., ``\textit{Mitochondria is the powerhouse of the cell.}'') are provided by the most expert model in the pool on a specific domain. Finally, the collected data are used with data-driven merging approaches~\cite{lmcocktail,evomerge}, where we optimize merging minimizing a loss on both alignment and domain-specific data. By doing this, we ensure that the merged model maintains high alignment and domain performance. \\\vspace{-10px}

\noindent Our contributions are threefold:
\begin{itemize}[noitemsep,topsep=0pt,leftmargin=*]
    \item We demonstrate that existing model merging techniques fail to explore the inherent trade-off between alignment and domain accuracy.
    \item We propose a safety-aware merging pipeline that achieves greater alignment of the merged model without sacrificing its accuracy.
    \item We present extensive experiments and ablations on the components of our pipeline, demonstrating its robustness in several conditions.
\end{itemize}
\section{Related work}\label{sec:related}

\paragraph{LLM Alignment} 
Ensuring the alignment of LLMs is crucial. 
Fine-tuning risks were highlighted by \citet{qi2023fine} and \citet{jain2024mechanistically}, showing that even benign datasets can degrade model safety and careful adaptation protocols are needed to preserve alignment.
Recently, some techniques to align LLM were proposed, such as ARGS~\cite{khanov2024args} addressing decoding, FIGA~\cite{guo2023beyond} for token-level signals, and f-DPO~\cite{wang2024beyond} for efficient alignment. \citet{zhao2023group} designed GPO to consider different interest groups. Some method enhance generalization~\cite{zheng2024improving}, while
\citet{dai2024safe} proposed Safe RLHF, for separate alignment on helpfulness and harmlessness.
In SALMON~\cite{sun2024salmon}, they use synthetic data to reduce human supervision. Although these may be effective, we show that model merging can mitigate the effects of alignment procedures.
Importantly, \citet{inan2023llama} addressed the need for effective input-output safeguarding in conversational AI with Llama Guard, employing a safety risk taxonomy and \textit{ad hoc} models to classify safety concerns in text.\looseness=-1

\paragraph{Model Merging} Techniques for merging multiple models have been proposed as efficient ways to benefit from the capabilities of multiple LLMs without retraining or accessing the original datasets. 
In Model Soups~\cite{wortsman2022model}, they first propose to combine models with weight averaging, showing improved performance compared to a single model. \citet{ilharco2022editing} build on this by performing task arithmetics, \ie element-wise operations on model parameters to edit their behavior towards specific tasks. Similar alternatives are RegMean \cite{regmean}, and Fisher Merging \cite{matena2022merging}. Model merging in non-linear spaces showed improved results, as in SLERP~\cite{slerp}. Some, such as TIES~\cite{ties} and DARE~\cite{dare}, propose methods to improve model merging, focusing on sparsification. Similarly, Model Breadcrumbs \cite{breadcrumbs} exploits sparse masks for better combination. Importantly, some extend merging capabilities across multiple modalities~\cite{sung2023empirical}. The importance of each model to merge can be automatically tuned with data-driven approaches such as EvoMM~\cite{evomerge} and LM-Cocktail~\cite{lmcocktail}. None of these approaches consider the safety implications of merging.

\paragraph{Alignment Evaluation} Advancements in evaluating LLMs have focused on their robustness, ethical considerations, and safety alignment. PromptBench \cite{zhu2023promptbench} offers a comprehensive benchmark to assess robustness against prompt perturbations, revealing vulnerabilities.
ReCode \cite{wang2023recode} proposes a similar setup for code generation.
\citet{ye2024flask} introduces FLASK, for a fine-grained assessment of alignment; while \citet{li2023generative} developed AUTO-J, a flexible generative judge. 
TrustGPT \cite{huang2023trustgpt} provides a benchmark for evaluating toxicity, bias, and value alignment. The ETHICS dataset \cite{hendrycks2020aligning} assesses understanding of ethics, while MoralChoice \cite{scherrer2024evaluating} analyzes moral beliefs in LLMs using psychological surveys and high ambiguity dilemmas. BeaverTails \cite{ji2024beavertails} introduces a dataset of over 700,000 questions and answers pairs annotated for helpfulness and harmlessness. Jailbreaking attacks' effectiveness is tackled in RigorLLM \cite{yuan2024rigorllm}. To the best of our knowledge, we are the first to evaluate the alignment of merged models.

\section{Preliminaries}\label{sec:preliminaries}
Here, we introduce notions and formalism on model merging. Merging aims to combine the specific capabilities of expert models, \ie, models finetuned on domain-specific data, into a single LLM.\looseness=-1

\subsection{Background on Model Merging}
Consider an ensemble of $N$ models $\mathcal{F}$. Each $\bm{f} \in \mathcal{F}$ is a model that excels in a specific domain, outperforming other models in domain-specific benchmarks. Let us define one $\bm{f}_{\text{base}} \in \mathcal{F}$ as the \textit{base} model, parameterized by $\bm{\theta}_{\text{base}} \in \mathbb{R}^d$. The choice of the base model is arbitrary. Similarly, the remaining $N-1$ expert models are defined as $\{\bm{f}^t_{\text{expert}}\}_{t=1}^{N-1}$, each parameterized by $\bm{\theta}_{\text{expert}}^t \in \mathbb{R}^d$.

Following~\citet{ilharco2022editing}, we define a task vector \(\bm{\tau}_t \in \mathbb{R}^d\) as the difference between the parameters of the expert and base models by
\begin{equation}    
\bm{\tau}_t = \bm{\theta}_{\text{expert}}^t - \bm{\theta}_{\text{base}}.
\end{equation}
We identify $\{\bm{\tau}_t\}_{t=1}^{N-1}$ as the set of task vectors. Using task arithmetic~\cite{ilharco2022editing}, a \textit{merged} model $\bm{f}_\text{merged}$ parameterized by \(\bm{\theta}_{\text{merged}} \in \mathbb{R}^d\) can be obtained, transferring the knowledge of multiple experts while preserving the expertise of the base model. This is generally written as:
\begin{equation}
    \bm{\theta}_{\text{merged}} = \bm{\theta}_\text{base} + \sum_{t=1}^{N-1}\lambda_t \bm{\tau}_t,
    \label{eqn:merge}
\end{equation}
where $\lambda_t \in \mathbb{R}$ are \textit{task weighting} factors that balance the performance on different tasks. Several approaches implement more advanced strategies for task vector combination, such as SLERP~\cite{slerp}, TIES~\cite{ties}, DARE~\cite{dare}, or DARE-TIES~\cite{dare,goddard2024arcee}. However, these still require manual tuning of the task weighting values $\lambda_t$, to balance the importance of each model during merging. %

\subsection{Automatic Task Weighting}\label{sec:evomerge}
The choice of $\lambda_t$ values significantly influences the effectiveness of existing merging techniques. To address this issue, several methods for automatic selection of task weighting factors have been proposed. For instance, \citet{evomerge} introduce EvoMM, an evolutionary-based algorithm for selecting the $\lambda_t$ using an iterative genetic algorithm such as CMA-ES~\cite{hansen2003reducing}. In each iteration, $\{\lambda_t\}_{t=1}^{N-1}$ values are randomly sampled $p$ times, where $p$ is a \textit{population} hyperparameter typical of genetic optimization~\cite{hansen2003reducing}. Assuming a merging algorithm like TIES~\cite{ties}, this generates $p$ different versions of $\bm{\theta}_\text{merged}$, which are then evaluated according to a user-defined criterion $\mathcal{C}$, such as accuracy on a downstream question-answering task evaluated on a set of datasets, for general or domain-specific knowledge evaluation. The goal of EvoMM is to find $\bm{\theta}_\text{merged}$ to maximize the performance, according to the criterion $\mathcal{C}$. The genetic algorithm assesses the effectiveness over the entire population of sampled $\bm{f}_\text{merged}$ on $\mathcal{C}$. In the next iteration, a new set of $\{\lambda_t\}_{t=1}^{N-1}$ are sampled close to the $\lambda_t$ resulting in the best-performing $\bm{f}_\text{merged}$. This process is repeated until convergence. In practice, \citet{evomerge} also use evolutionary algorithms to optimize method-specific hyperparameters for SLERP~\cite{slerp}, TIES~\cite{ties}, and DARE~\cite{dare}.

Alternatively, LM-Cocktail~\cite{lmcocktail} proposes a method for identifying $\lambda_t$ based on performance on a few samples. Assuming a dataset $\mathcal{D}$ composed of a few domain-specific questions and answers $(q, a)$, they design a heuristic that balances the contributions of existing models based on their performance on $\mathcal{D}$. This is formulated as:
\begin{equation}\label{eqn:lmcocktail}
    \begin{split}
    w_t = \mathbb{E}_{(q, a)\sim\mathcal{D}}[-\mathcal{L}_\text{ce}(\bm{f}^t_{\text{expert}}(q), a)],    \\
    \{\lambda_t\}_{t=1}^{N-1} = \text{softmax}(\{w_t\}_{t=1}^{N-1}),
\end{split}
\end{equation}
where $\mathcal{L}_\text{ce}$ refers to the cross-entropy loss between the model prediction and the ground-truth answer. In LM-Cocktail, $\{\lambda_t\}_{t=1}^{N-1}$ are the terms of a linear combination of weights $\{\bm{\theta}_t\}_{t=1}^{N-1}$ rather than of task vectors $\{\tau_t\}_{t=1}^{N-1}$. For more details, we refer to~\citet{lmcocktail}. A common aspect of both approaches for automatic task weighting is the usage of external data. In the next section, we exploit this characteristic to enforce safety alignment in merged models while maximizing accuracy.

\begin{figure}[t]
    \centering
    \includegraphics[width=\linewidth]{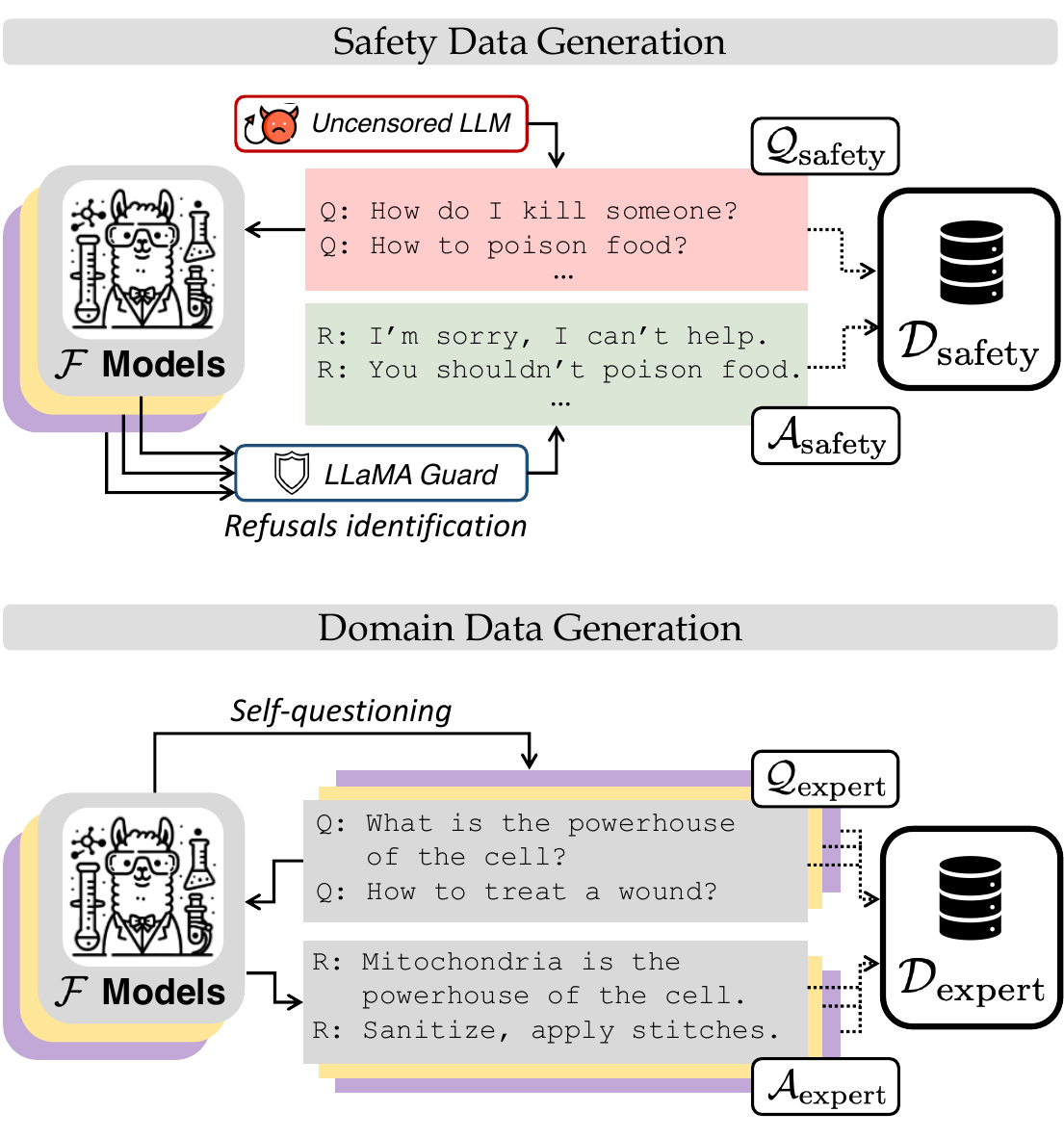}
    \caption{\textbf{Data generation.} We generate both safety data $\mathcal{D}_\text{safety}$ (top) and expert domain data $\mathcal{D}_\text{expert}$ (bottom). For safety data, we use an uncensored LLM to generate harmful questions, and collect refusals of the $\mathcal{F}$ experts with LLaMA-Guard~\cite{metallamaguard2}. For domain data, we use the $\mathcal{F}$ experts to generate questions in different domains (\textit{self-questioning}) and collect responses.\looseness=-1}
    \label{fig:data}
\end{figure}

\section{Safety-Aware Merging}\label{sec:method}

\subsection{Motivation}\label{sec:method-motivation}
We recall that although merging techniques are effective for boosting performance on downstream datasets~\cite{slerp,ties,dare}, an important aspect has been overlooked in the literature: \textit{there is no consideration of safety alignment in the merging process}. Naively merging models with existing techniques can result in the removal of safety alignment, as shown later in Section \ref{sec:exp}. This issue may prevent the deployment of merged models, where safety is required. In this section, we build on state-of-the-art data-dependent automatic task weighting strategies~\cite{evomerge,lmcocktail} to propose simple baselines for safety-aware merging.\looseness=-1

Our intuition is that safety alignment should be treated \textit{as a task in its own right}. Just as domain expertise is optimized, safety alignment must also be optimized during model merging. Current automatic task weighting methods rely on data to optimize performance and to achieve our goal, we need to incorporate both alignment data and domain data into the optimization process. 
By leveraging this data dependency, we can ensure that the merged model retains both domain expertise and safety alignment incorporated in the data. Moreover, we propose a fully automated pipeline, relying on synthetic data only. While we still retain compatibility with public datasets, this allows us to avoid external dependencies in the merging process. Next, we describe our data generation pipeline.

\subsection{Safety Data Generation}\label{sec:method-safety-data-generation}
As introduced in Section~\ref{sec:intro}, the goal of safety alignment in LLMs is to respond to \textit{unsafe} input prompts with refusals, \ie, sentences like “\textit{I am sorry, but I cannot help}”. This is typically achieved through fine-tuning on unsafe prompts and their corresponding refusals~\cite{ouyang2022training}. However, models in the merging set $\mathcal{F}$ may have been trained with different data and procedures, leading to varying levels of safety alignment. Therefore, it is important that the merged model $\bm{f}_\text{merged}$ reproduces the refusals of models $\bm{f}\in\mathcal{F}$ for unsafe inputs.\\\vspace{-6px}

We start by generating a set of $K$ unsafe questions $\mathcal{Q}_\text{safety}$. We use an uncensored LLM\footnote{\url{https://huggingface.co/cognitivecomputations/dolphin-2.9-llama3-8b}} to generate $\mathcal{Q}_\text{safety}$, since safety-aligned LLMs in $\mathcal{F}$ may refuse to generate such questions. Details of our prompt are provided in Appendix~\ref{sec:appendix-data-generation}. This can be replaced with pre-generated unsafe inputs from datasets such as BeaverTails~\cite{ji2024beavertails}. We then use $q_\text{safety}\sim\mathcal{Q}_\text{safety}$ as input for all $\bm{f}\in\mathcal{F}$, collecting a set of replies for each prompt $q_\text{safety}$. These replies are processed with LLaMA-Guard 2~\cite{metallamaguard2} to identify refusals. We randomly select one refusal $a_\text{safety}$ for each $q_\text{safety}$. By repeating this for all $q_\text{safety} \in \mathcal{Q}_\text{safety}$, we obtain a set of refusals $\mathcal{A}_\text{safety}$. This results in a safety dataset of unsafe questions and associated refusals, $\mathcal{D}_\text{safety} = \{(q^i_\text{safety}, a^i_\text{safety})\}_{i=1}^K$, where $q^i_\text{safety}\in\mathcal{Q}_\text{safety},a^i_\text{safety}\in\mathcal{A}_\text{safety}$. The process is shown in Fig.~\ref{fig:data} (top). If no model in $\mathcal{F}$ replies with a refusal, the input $q_\text{safety}$ is discarded. 

\subsection{Domain Data Generation}\label{sec:method-task-data-generation}
Besides preserving alignment, we aim to transfer the expertise of each $\bm{f}_\text{expert}^t$ to $\bm{f}_\text{merged}$. To do this, we generate a Q\&A dataset for each domain of expertise to optimize task weighting.

We use the expert models to generate questions. Each $\bm{f}_\text{expert}^t$ is prompted to generate an expert-specific question $q_\text{expert}^{t}$. For instance, if $\bm{f}_\text{expert}^t$ specializes in mathematics, we will use it to generate math-related questions. We use in-context learning to provide examples of questions. Then, we prompt $\bm{f}_\text{expert}^t$ with $q_\text{expert}^{t}$ to obtain a corresponding answer $a_\text{expert}^{t}$. This \textit{self-questioning} procedure is inspired by related literature~\cite{li2022self,press2022measuring}. Each model $\bm{f}_\text{expert}^t$ produces $K/(N-1)$ questions and associated answers, hence we can aggregate all questions and answers in two sets $\mathcal{Q}_\text{expert}$ and $\mathcal{A}_\text{expert}$, respectively, both of size $K$. Finally, we construct $\mathcal{D}_\text{expert} = \{(q^i_\text{expert}, a^i_\text{expert})\}_{i=1}^K$, where $q^i_\text{expert}\in\mathcal{Q}_\text{expert},a^i_\text{expert}\in\mathcal{A}_\text{expert}$. This process is shown in Fig.~\ref{fig:data} (bottom). 
Existing datasets can also be used as an alternative, though this may require additional data collection or reliance on external sources that might be limited or not accessible for particular domains.

\subsection{Merging}\label{sec:method-merging}
We use the previously collected datasets, $\mathcal{D}_{\text{safety}}$ and $\mathcal{D}_{\text{expert}}$, to guide the optimization of task weights $\lambda_t$, maximizing both alignment and domain performance. By leveraging automatic task weighting strategies that depend on data, such as EvoMM~\cite{evomerge} and LM-Cocktail~\cite{lmcocktail}, we ensure that the merged model retains both safety alignment and domain expertise. We propose a custom safety-aware adaptation of both EvoMM and LM-Cocktail.\looseness-1

\vspace{7px}
For EvoMM, we optimize the merged model $\bm{f}_\text{merged}$ to output an associated response $a$, given a question $q$, where the pair ($q$,$a$) is sampled from either $\mathcal{D}_\text{safety}$ or $\mathcal{D}_\text{expert}$. This ensures that the resulting $\bm{f}_\text{merged}$ preserves both the safety alignment of existing models in $\mathcal{F}$ and their expertise in various domains. Formally, given ${(q_\text{safety}, a_\text{safety})\sim\mathcal{D}_\text{safety}}$ and ${(q_\text{expert}, a_\text{expert})\sim\mathcal{D}_\text{expert}}$, we impose a cross-entropy loss $\mathcal{L}_\text{ce}$ between the answer generated by $\bm{f}_\text{merged}(q)$ and the associated reply $a$. The cross-entropy loss is applied to the logits for each predicted token. We formulated it as:
\begin{equation}
\begin{split}
\mathcal{L}_r = \mathbb{E}_{(q_r, a_r)\sim\mathcal{D}_r}[-\mathcal{L}_\text{ce}(\bm{f}_\text{merged}(q_r), a_r)],\\
r \in \{\text{safety, expert}\}.
\end{split}
\end{equation}
We combine the two terms into a single loss, using a factor $\alpha$ to balance each contribution by
\begin{equation}\label{eqn:merge-loss}
    \mathcal{L}_\text{merge} = \mathcal{L}_\text{safety} + \alpha \mathcal{L}_\text{expert}.
\end{equation}
We then assume $\mathcal{C}=\mathcal{L}_\text{merge}$ and optimize over $\{\lambda_t\}_{t=1}^{N-1}$. In other words, we use the merged model $\bm{f}_\text{merged}$ to process both $\mathcal{D}_\text{expert}$ and $\mathcal{D}_\text{safety}$, optimizing $\{\lambda_t\}_{t=1}^{N-1}$ to maximize performance on both. We recall that $\bm{\theta}_\text{merged}$ is obtained with Eq.~\eqref{eqn:merge}.\looseness=-1

For LM-Cocktail~\cite{lmcocktail}, instead, we assume $\mathcal{D} = \mathcal{D}_\text{safety} \cup \mathcal{D}_\text{expert}$, and calculate $\{\lambda_t\}_{t=1}^{N-1}$ applying Eq.~\eqref{eqn:lmcocktail} for all $\{\bm{f}_\text{expert}^t\}_{t=1}^{N-1}$.

\begin{table*}[t]
  \centering
  \resizebox{\linewidth}{!}{%
  \begin{tabular}{cll|ll|ll}
  & &\multicolumn{1}{c}{} & \multicolumn{2}{c}{$\mathcal{F}=\text{\{\textit{Mistral, MAmmoTH}}\}$} & \multicolumn{2}{c}{$\mathcal{F}=\text{\{\textit{Llama, OpenBioLLM}}\}$}\\
  \toprule
   \multirow{2}{*}{\textbf{Merging}}&\multirow{2}{*}{\textbf{Task Weighting}}& \multicolumn{1}{c}{\multirow{2}{*}{\textbf{Data}}} & \multicolumn{1}{c}{\multirow{2}{*}{\textbf{Alignment~$\uparrow$}}} & \multicolumn{1}{c}{\textbf{Accuracy~$\uparrow$}} & \multicolumn{1}{c}{\multirow{2}{*}{\textbf{Alignment~$\uparrow$}}} & \multicolumn{1}{c}{\textbf{Accuracy~$\uparrow$}}\\
   && \multicolumn{1}{c}{}& \multicolumn{1}{c}{} & \multicolumn{1}{c}{\small{(\textit{STEM})}} & & \multicolumn{1}{c}{\small{(\textit{BIO})}}\\
  \midrule
  \multicolumn{3}{c|}{\textit{Expert models $\mathcal{F}$}} & \textit{91.5 / 64.8} & \textit{49.6 / 53.1} & \textit{97.9 / 48.3} & \textit{68.9 / 71.8} \\
  \midrule\midrule
\multirow{3}[0]{*}{TIES} & Grid search  & -  & 72.7    & 53.7  & 89.3 & \textbf{74.1}  \\
      & EvoMM & $\mathcal{D}_\text{expert}$     & 61.6 \diff{-11.1}  & 52.0 \diff{-1.7} & 79.8 \diff{-9.5} & 73.2 \diff{-0.9} \\
      & EvoMM (ours) & $\mathcal{D}_\text{expert} \cup \mathcal{D}_\text{safety}$      & \textbf{78.1} {\diff{+5.4}}     & \textbf{54.2} \diff{+0.5}  & \textbf{96.0} \diff{+6.7} & 73.6 \diff{-0.5} \\\midrule
\multirow{3}[0]{*}{DARE-TIES} & Grid search  & -     & 72.9    & 53.3  & 89.3 & \textbf{74.1} \\
      & EvoMM & $\mathcal{D}_\text{expert}$     & 60.6 \diff{-12.3}  & 51.7 \diff{-1.6} & 80.1 \diff{-9.2} & 73.8 \diff{-0.3} \\
      & EvoMM (ours) & $\mathcal{D}_\text{expert} \cup \mathcal{D}_\text{safety}$    & \textbf{78.3} \diff{+5.4}   & \textbf{54.0} \diff{+0.7} & \textbf{96.1} \diff{+6.8} & 73.8 \diff{-0.3} \\\midrule
\multirow{3}[0]{*}{SLERP} & Grid search  & -     & 75.1     & 53.7 & 82.3 & 74.1 \\
      & EvoMM & $\mathcal{D}_\text{expert}$     & 71.9 \diff{-3.2}  & 52.9 \diff{-0.8} & 86.0 \diff{+3.7} & \textbf{74.2} \diff{+0.1} \\
      & EvoMM (ours) & $\mathcal{D}_\text{expert} \cup \mathcal{D}_\text{safety}$    & \textbf{77.6} \diff{+2.5}    & \textbf{54.0} \diff{+0.3} & \textbf{90.7} \diff{+8.4} & \textbf{74.2 }\diff{+0.1} \\\midrule\midrule
      \multirow{2}{*}{-}&\multicolumn{1}{l}{LM-Cocktail} & $\mathcal{D}_\text{expert}$     & 72.5       & \textbf{53.3} & 92.6 & \textbf{74.1} \\
    &\multicolumn{1}{l}{LM-Cocktail (ours)} & $\mathcal{D}_\text{expert} \cup \mathcal{D}_\text{safety}$      & \textbf{74.4} \diff{+1.9}   & 53.2 \diff{-0.1} & \textbf{94.1} \diff{+1.5} & 74.0 \diff{-0.1} \\\bottomrule
  \end{tabular}}
\caption{\textbf{Benchmark of safety-aware merging.} We report performance in two different $\mathcal{F}$ setups, achieving aligned models expert in STEM and biology. We compare with baselines performing manual hyperparameter search (grid search) or using automatic task weighting strategies with $\mathcal{D}_\text{expert}$ only. Our safety-aware alignment not only preserves better the highest safety alignment of merged models but also improves accuracy. Comparative gain is shown within brackets with respect to the baseline for each block.}
\label{tab:main_results}
\end{table*}
\section{Experiments}\label{sec:exp}

\subsection{Experimental Setup}\label{sec:setup}

\noindent\textbf{Merging Techniques\ } We use two automatic methods to find the task weights of Eq.~\eqref{eqn:merge}, \ie, EvoMM~\cite{evomerge} and LM-Cocktail~\cite{lmcocktail}, in which we add safety alignment data following Section~\ref{sec:method-merging}. As recommended~\cite{evomerge}, we use EvoMM for optimization on top of DARE-TIES~\cite{dare,goddard2024arcee}, and we add TIES~\cite{ties} and SLERP~\cite{slerp} as merging algorithm for completeness. For all, we report the merged models maximizing domain accuracy. We use MergeKit \cite{goddard2024arcee} as codebase. More details are in Appendix~\ref{sec:appendix-implementation-details}.\\

\noindent\textbf{Models \ } We use five LLMs for our experiments, \ie Mistral-0.2-7B-Instruct~\cite{mistral}, LLaMA-3-8B-Instruct~\cite{llama3modelcard}, OpenBioLLM-8B~\cite{openbio}, MAmmoTH-2-7B~\cite{mammoth}, and WizardMath-1.1-7B~\cite{luo2023wizardmath} - in the following we drop versions for brevity. Among them, we consider experts in the biology (OpenBioLLM), STEM (MAmmoTH), and math (WizardMath) domains, as well as instruction-finetuned models (Mistral, LLaMA). We set general-purpose models (Mistral, LLaMA) as $\bm{f}_\text{base}$. Note that although these models lack domain expertise, they are finetuned on safety instructions for refusals generation; hence, \textit{they exhibit safety properties that we are interested in preserving}. For each expert, we generate domain data $\mathcal{D}_\text{expert}$ following the self-questioning procedure introduced in Section~\ref{sec:method-task-data-generation} with custom prompts, capturing specific expertise. We report prompts for data generation in Appendix~\ref{sec:appendix-data-generation}.\\

\noindent\textbf{Evaluation\ }
To evaluate alignment, we use the BeaverTails30K~\cite{ji2024beavertails} test set, including 1,733 unsafe prompts, for which aligned language models are expected to generate refusals. We generate responses for each prompt with our obtained models and use
LLaMA-Guard-2 \cite{metallamaguard2} for flagging the answers as safe or unsafe. Finally, we report the percentage of safe outputs (\ie, refusals) as an alignment metric. For domain performance, we use specific benchmarks related to domain expertise. We consider a \textit{STEM} set composed of some STEM subjects from MMLU \cite{hendrycks2020measuring} as defined in \cite{azerbayev2023llemma}; a
\textit{BIO} set, composed of MedMCQA \cite{pal2022medmcqa}, MedQA-USMLE-4-options \cite{jin2021disease}, PubMedQA \cite{jin2019pubmedqa}, and six biology-related subjects from MMLU: College Biology, College Medicine, Anatomy, Pro Medicine, Medical Genetics, and Clinical KG~\cite{hendrycks2020measuring}. We also use the commonsense reasoning WinoGrande~\cite{sakaguchi2021winogrande} and the science-related reasoning ARC~\cite{clark2018think} datasets.
For each benchmark, we calculate the model accuracy on multiple choice or binary classification tasks with LM Harness \cite{lmharness}.\looseness=-1

\subsection{Safety-Aware Merging Performance}\label{sec:main_results}

\paragraph{Benchmark}
In Table~\ref{tab:main_results} we present results across merging configurations with $N=2$. We aim to obtain merged models with good domain expertise and desirable safety alignment. First, we consider $\mathcal{F}=\{\text{Mistral, MAmmoTH}\}$, %
to obtain an aligned STEM expert. Here, we evaluate performance on the \textit{STEM} set. In a second set of experiments, we consider $\mathcal{F}=\{\text{LLaMA, OpenBioLLM}\}$,
to get an aligned biology expert. For the latter, we evaluate the accuracy on the \textit{BIO} set.
We report the average accuracy across all datasets in the splits.%

\noindent We first verify performance of the models in $\mathcal{F}$ for both setups. In Table~\ref{tab:main_results}, first row, we show that base models are most aligned, with \textbf{91.5} alignment for Mistral and \textbf{97.9} for LLaMA. Expert models report better performance in domain-specific tasks, such as \textbf{53.1} for MAmmoTH on \textit{STEM} (\textit{vs} \textbf{49.6} for Mistral) and \textbf{71.8} for OpenBioLLM on \textit{BIO} (\textit{vs} \textbf{68.9} for LLaMA), while they both lack safety alignment (\textbf{64.8} and \textbf{48.3}, respectively).\\[-0.1in]

\begin{table}[t]   
    \setlength{\tabcolsep}{2pt}
    \centering
    \resizebox{\linewidth}{!}{
    \begin{tabular}{cll|ccc}
    \multicolumn{6}{c}{$\mathcal{F}=\text{\{\textit{Mistral, WizardMath, MAmmoTH}}\}$}\\
        \toprule
  \multirow{2}{*}{} & \multicolumn{1}{l}{\textbf{\multirow{2}{*}{Task Weight.}}} & \multicolumn{1}{c}{\multirow{2}{*}{\textbf{Data}}} & \multicolumn{1}{c}{\multirow{2}{*}{\textbf{Align.~$\uparrow$}}} & \multicolumn{2}{c}{\textbf{Acc.~$\uparrow$}}\\
  &&\multicolumn{1}{c}{}&& \small{(ARC)} & \small{(WG)} \\\midrule
  \multicolumn{3}{c|}{{\textit{Mistral}}} & \multicolumn{1}{c}{\textit{91.6}} & \multicolumn{1}{c}{\textit{54.4}} & \multicolumn{1}{c}{\textit{73.5}} \\
  \multicolumn{3}{c|}{{\textit{WizardMath}}} & \multicolumn{1}{c}{\textit{80.0}} & \multicolumn{1}{c}{\textit{51.3}} & \multicolumn{1}{c}{\textit{74.2}} \\
  \multicolumn{3}{c|}{{\textit{MAmmoTH}}} & \multicolumn{1}{c}{\textit{65.0}} & \multicolumn{1}{c}{\textit{57.2}} & \multicolumn{1}{c}{\textit{70.6}} \\\midrule\midrule
  \multirow{3}{*}{\rotatebox{90}{TIES}} & Grid Search & - & 76.5& 59.2 & 74.8\\
  & EvoMM & $\mathcal{D}_\text{expert}$ & 49.1 & 56.2 & 73.2 \\
  & EvoMM (ours) & $\mathcal{D}_\text{expert} \cup \mathcal{D}_\text{safety}$ & \textbf{76.6} & \textbf{59.6} & \textbf{75.2} \\\midrule
  \multirow{3}{*}{\rotatebox{90}{DARE-T.}}  & Grid Search & - & 72.8 & 59.4 & \textbf{74.6} \\
  & EvoMM & $\mathcal{D}_\text{expert}$ & 48.3 & 57.0 & 74.0\\
  & EvoMM (ours) & $\mathcal{D}_\text{expert} \cup \mathcal{D}_\text{safety}$ & \textbf{81.1} & \textbf{59.6} & 73.7 \\\midrule\midrule
      \multirow{2}{*}{}
  & LM-Cocktail & $\mathcal{D}_\text{expert}$ & 62.3 & 58.0 & 73.6\\
  & LM-Cocktail (ours) & $\mathcal{D}_\text{expert} \cup \mathcal{D}_\text{safety}$ & \textbf{65.3} & \textbf{58.3} & \textbf{74.1} \\\midrule
    \end{tabular}}
    \caption{\textbf{Merging three models.} Benchmarks of three models and their merged counterparts. With the addition of $\mathcal{D}_\text{safety}$, we considerably increase both alignment and domain accuracy on WinoGrande (WG) and ARC, for both EvoMM and LM-Cocktail.}
    \label{tab:3way_merging}
\end{table}

\noindent We then propose strong \textit{grid search} baselines, by extensively optimizing manually task weights and hyperparameters for the TIES, DARE-TIES, and SLERP merging algorithms. These baselines do not use auxiliary data for the optimization of task weights, but they requires considerable computation times due to the multiple configurations available. Then, we present results for the data-driven strategies EvoMM and LM-Cocktail using $\mathcal{D}_\text{expert}$ only. We use EvoMM to optimize $\{\lambda_t\}_{t=1}^{N-1}$ and hyperparameters of the task vector combination algorithm (\ie, TIES, DARE-TIES, and SLERP), as we detail in Appendix~\ref{sec:appendix-implementation-details}. For LM-Cocktail, we follow~\cite{lmcocktail} and optimize $\{\lambda_t\}_{t=1}^{N-1}$ only.\looseness=-1

\noindent We finally report our safety-aware merging performance, by including $\mathcal{D}_\text{safety}$ in each data-driven merging strategy. For EvoMM, we show that including safety data achieves the highest alignment of merged models, reporting for instance \textbf{96.1} in DARE-TIES with EvoMM, only \textbf{1.8} below the original LLaMA (\textbf{98.0}), while EvoMM using only $\mathcal{D}_\text{expert}$ falls short at \textbf{80.1}. Also, we highlight how we achieve great accuracy across all setups, always outperforming single experts in $\mathcal{F}$ and, in many scenarios, \textit{even outperforming corresponding safety-unaware baselines}. Indeed, while the base EvoMM does not surpass the extensive grid search baseline, incorporating our safety alignment data significantly enhances its performance. This may be caused by the usage of data beyond $\mathcal{D}_\text{expert}$, that help regularize the optimization process, converging to better minima for $\{\lambda_t\}_{t=1}^{N-1}$. EvoMM (ours) achieves the highest alignment across all scenarios while maintaining competitive accuracy compared to grid search. Our results are consistent using LM-Cocktail too, where we improve alignment in both scenarios (\textcolor{OliveGreen}{\textbf{+1.9}} and \textcolor{OliveGreen}{\textbf{+1.5}}, respectively) while achieving on-par domain accuracy compared to the baseline LM-Cocktail with only $\mathcal{D}_\text{expert}$.

\paragraph{Merging Beyond Two Models}

We investigate the potential of safety-aware merging with a pool of experts $\mathcal{F}$ encompassing more than two models. In this setup, we consider $\mathcal{F}$ composed by: Mistral, MAmmoTH, and WizardMath. We specifically design this setup since, although both MAmmoTH and WizardMath are finetuned on similar domains, they exhibit significant differences in performance on the Winogrande and ARC benchmarks, as empirically verified in Table~\ref{tab:3way_merging}. Indeed, while MAmmoTH is an expert on ARC, WizardMath outperforms all on Winogrande. Mistral is an expert in alignment, reporting \textbf{91.6} on BeaverTails30K.\\

We report results following our setup in Table~\ref{tab:main_results}.
Note that SLERP is not applicable since it is only usable when $N=2$. Table~\ref{tab:3way_merging} shows that our safety-aware merging achieves the highest alignment across all scenarios. Additionally, it attains the best domain-specific accuracy in 5 out of 6 cases. Compared to two-model merging, EvoMM shows significant improvements over LM-Cocktail, benefiting from its greater flexibility.

\subsection{Ablation studies}
\label{sec:experiments-ablation}

\begin{figure}[t]
    \centering
    \includegraphics[width=\linewidth]{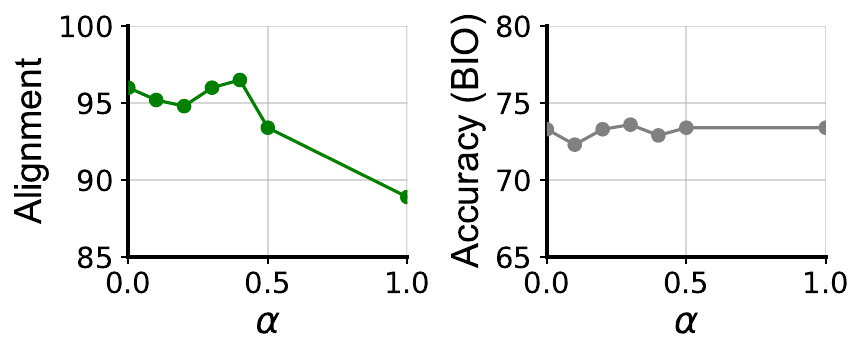}  \vspace{-25px}
    \caption{\textbf{Varying loss combination factor $\alpha$.} For $\alpha\le0.5$, merging yields good results in both accuracy and alignment. For greater $\alpha$ (\eg 1.0), alignment degrades significantly while accuracy does not improve.\looseness=-1}
    \label{fig:plot_alpha}
\end{figure}
\begin{table}[t]
  \centering
  \begin{subtable}{\linewidth}
  \setlength{\tabcolsep}{0.02\linewidth}
  
  \resizebox{\linewidth}{!}{%
    \begin{tabular}{cllc|cc}
    \multicolumn{6}{c}{$\mathcal{F}=\text{\{\textit{LLaMA, OpenBioLLM}}\}$}\\
    \toprule

    & \multicolumn{1}{c}{\textbf{\multirow{2}{*}{Task Weight.}}} & \multicolumn{1}{c}{\multirow{2}{*}{\textbf{Data}}} & \multicolumn{1}{c}{\multirow{2}{*}{\textbf{Real}}} & \multicolumn{1}{c}{\multirow{2}{*}{\textbf{Align.~$\uparrow$}}} & \multicolumn{1}{c}{\textbf{Acc.~$\uparrow$}}\\
    & & & \multicolumn{1}{c}{}& & (\textit{BIO})\\
    \midrule
    \multicolumn{4}{c|}{\textit{LLaMA / OpenBioLLM}} & \textit{97.9 / 48.3} & \textit{68.9 / 71.8}\\\midrule\midrule
    \multirow{4}{*}{\rotatebox{90}{TIES}} & \multirow{2}{*}{EvoMM} & \multirow{2}{*}{$\mathcal{D}_\text{expert}$} & \xmark & 79.8 & 73.2 \\
    & & & \checkmark & \textbf{89.7} & \textbf{73.8}\\\cmidrule{2-6}
    & \multirow{2}{*}{EvoMM (ours)} & \multirow{2}{*}{$\mathcal{D}_\text{expert}\cup\mathcal{D}_\text{safety}$} & \xmark & 96.0 & 73.6 \\
    & & & \checkmark & \textbf{96.2} &	\textbf{73.8}\\\midrule
    
    \end{tabular}%
    }
    \caption{Generated data vs.\ Real data}\label{tab:ablation_data_source}%
      \vspace{10px}

  \end{subtable}
  \begin{subtable}{\linewidth}
      \setlength{\tabcolsep}{0.02\linewidth}
  
  \resizebox{\linewidth}{!}{%
    \begin{tabular}{cccc|cc}
    \multicolumn{6}{c}{$\mathcal{F}=\text{\{\textit{LLaMA, OpenBioLLM}}\}$}\\

    \toprule

    & \multicolumn{1}{c}{\textbf{\multirow{2}{*}{Task Weight.}}} & \multicolumn{1}{c}{\multirow{2}{*}{\textbf{Data}}} & \multicolumn{1}{c}{\multirow{2}{*}{\textbf{$K$}}} & \multicolumn{1}{c}{\multirow{2}{*}{\textbf{Align.~$\uparrow$}}} & \multicolumn{1}{c}{\textbf{Acc.~$\uparrow$}}\\
    & & & \multicolumn{1}{c}{}& & (\textit{BIO})\\
    \midrule
    \multicolumn{4}{c|}{\textit{LLaMA / OpenBioLLM}} & \textit{97.9 / 48.3} & \textit{68.9 / 71.8}\\\midrule\midrule
    \multirow{3}{*}{\rotatebox{90}{TIES}} & \multirow{3}{*}{EvoMM (ours)} & \multirow{3}{*}{$\mathcal{D}_\text{expert}\cup\mathcal{D}_\text{safety}$} & 200 & 95.6 & 73.7 \\
    
    &&& 500 & 93.0 & \textbf{73.8}\\
    &&& 1000 & \textbf{96.0} & 73.6\\
    \bottomrule
    
    \end{tabular}%
    }
    \caption{Importance of $K$}\label{tab:k-effects}%

  \end{subtable}
  \caption{\textbf{Effects of data in the LLaMA-OpenBioLLM merge.} Table~(\subref{tab:ablation_data_source}) shows how replacing our data generation pipeline with real data sampled from the validation set of the target domain data results only in a minor performance increase. Table~(\subref{tab:k-effects}) ablates the effect of $K$, \ie, the number of samples in $\mathcal{D}_\text{expert}$ and $\mathcal{D}_\text{safety}$.}
\end{table}

In this section, we present ablation studies. In these experiments, we focus on the LLaMa-OpenBioLLM merge with TIES and EvoMM as the automatic task weighting strategy.

\paragraph{Impact of $\alpha$}
In Section~\ref{sec:method-merging}, we introduce $\alpha$, used to balance the importance of the two loss terms $\mathcal{L}_\text{safety}$ and $\mathcal{L}_\text{expert}$ in $\mathcal{L}_\text{merge}$ (Eq.~\eqref{eqn:merge-loss}). We test our safety-aware setup with different values of $\alpha$ in Fig.~\ref{fig:plot_alpha}. We highlight that for $\alpha\le0.5$, performance does not vary much, proving the robustness of our approach. Interestingly, even with $\alpha=0$, equivalent to using $\mathcal{D}_\text{safety}$ data only, performance remains competitive in accuracy. This shows that safety data may sometimes be sufficient to drive the merging procedure towards an acceptable combination of $\{\lambda_t\}_{t=1}^{N-1}$.
We choose $\alpha=0.3$ as the value maximizing the accuracy and use it for our experiments, yielding \textbf{73.6} accuracy and \textbf{96.0} alignment.
Higher $\alpha$ (\eg, $\alpha=1$) leads to saturation of the accuracy (\textbf{73.4}), but at a great cost for alignment (\textbf{88.9}). \looseness=-1

\paragraph{Data Source}
In Sections~\ref{sec:method-safety-data-generation} and~\ref{sec:method-task-data-generation}, we describe how to generate $\mathcal{D}_{\text{safety}}$ and $\mathcal{D}_{\text{expert}}$ using models in $\mathcal{F}$, hence avoiding to rely on external data. Here we test performance with \textit{real} data, constructing $\mathcal{D}_\text{expert}$ and $\mathcal{D}_\text{safety}$ by collecting samples from the validation set of existing benchmarks. We collect $K=1000$ prompts from the \textit{BIO} validation set (see Section~\ref{sec:setup}), and $K=1000$ instances from BeaverTails30K training set \cite{ji2024beavertails}. We then follow Section~\ref{sec:method} to generate responses to the collected questions. Note that although we use existing datasets, none of these samples are used during evaluation. We show results in Table~\ref{tab:ablation_data_source}. 
Real data significantly benefits the baseline EvoMM, improving accuracy by \textbf{\diff{+0.6}} and alignment by \textbf{\diff{+9.9}}. In contrast, our safety-aware pipeline shows minimal gains \textbf{\diff{+0.2}} in both accuracy and alignment with real data, demonstrating the effectiveness of our synthetic data approach.
When using real data, both methods achieve comparable accuracy, but our safety-aware EvoMM maintains a substantially higher alignment \textbf{\diff{+6.5}}.

\paragraph{Number of Samples}
Safety-aware merging requires $K$ samples in each $\mathcal{D}_\text{expert}$ and $\mathcal{D}_\text{safety}$. We study the importance of $K$ in Table~\ref{tab:k-effects}, showing results for $K\in\{200, 500, 1000\}$.
We report that accuracy is marginally impacted by increasing $K$, while alignment is more heavily influenced, achieving \textbf{96.0} alignment for $K=1000$, where the second best value is \textbf{95.6} for $K=200$. 
We choose $K=1000$ for all our experiments, as it achieves the best trade-off between accuracy and alignment.\looseness=-1

\begin{table}[t]
  \centering
  \setlength{\tabcolsep}{0.02\linewidth}
  
  \resizebox{\linewidth}{!}{%
    \begin{tabular}{cccc|cc}
    \multicolumn{6}{c}{$\mathcal{F}=\text{\{\textit{LLaMA, OpenBioLLM}}\}$}\\

    \toprule

    & \multicolumn{1}{c}{\textbf{\multirow{2}{*}{Task Weight.}}} & \multicolumn{1}{c}{\multirow{2}{*}{\textbf{Data}}} & \multicolumn{1}{c}{\multirow{2}{*}{\textbf{Steps}}} & \multicolumn{1}{c}{\multirow{2}{*}{\textbf{Align.~$\uparrow$}}} & \multicolumn{1}{c}{\textbf{Acc.~$\uparrow$}}\\
    & & & \multicolumn{1}{c}{}& & (\textit{BIO})\\
    \midrule
    \multicolumn{4}{c|}{\textit{LLaMA / OpenBioLLM}} & \textit{97.9 / 48.3} & \textit{68.9 / 71.8}\\\midrule\midrule
    \multirow{4}{*}{\rotatebox{90}{TIES}} & \multirow{4}{*}{EvoMM (ours)} & \multirow{4}{*}{$\mathcal{D}_\text{expert}\cup\mathcal{D}_\text{safety}$} & 50 & 95.7 & 73.2 \\
    &&& 100 & 96.0 & \textbf{73.6}\\
    &&& 200 & \textbf{97.3} & 72.2\\
    &&& 300 & 96.3 & 72.8\\
    \bottomrule
    \end{tabular}%
    }
    \caption{\textbf{Optimization steps for EvoMM.} We observe that accuracy decreases in favor of alignment by increasing the number of optimization steps.\looseness=-1}\label{tab:opt_steps}%

\end{table}

\paragraph{Optimization Steps for EvoMM}
Evolutionary optimization algorithms such as CMA-ES~\cite{hansen2003reducing} are iterative in nature. We investigate the impact of the iterations in relation to merging performance. In Table~\ref{tab:opt_steps}, we vary the optimization steps in EvoMM. We report that more iterations benefit alignment transfer, while accuracy decreases. We attribute this behavior to the greater difficulty of the alignment task, requiring more steps to be effectively transferred in $\bm{f}_\text{merged}$. Due to the increased optimization times for more steps, we perform our experiments with 100 steps, guaranteeing the best trade-off between performance and optimization times.\looseness=-1

\section{Conclusions}\label{sec:conclusions}
In this work, we have highlighted the effects of model merging in the context of safety alignment for LLMs. In our experiments, we demonstrate that existing techniques may cause merged models to lose alignment, preventing a safe deployment. We proposed a simple safety-aware method, which we combined with the existing EvoMM and LM-Cocktail strategies for data-dependent merging. By treating alignment as a task in its own right and incorporating alignment data into the merging process, our safety-aware merging pipeline significantly improves alignment, without compromising domain accuracy.

\section{Limitations}

Our work represents an initial exploration into the important issue of safety alignment in model merging. To the best of our knowledge, we are the first to explore such a setup. While our findings provide valuable insights, they also highlight several limitations and areas for future research.

\paragraph{Alignment Requirements} A key assumption in our approach is that at least one model in the merging pool is sufficiently aligned. This prerequisite may not always be met, especially when working with LLMs that have been trained on uncensored data. More in general, we showcase how performance depends on the alignment performance of the best model in $\mathcal{F}$, as evident in Table~\ref{tab:main_results}. We recommend always assessing the alignment of all models in $\mathcal{F}$ before merging. Moreover, future work should investigate methods to perform safety-aware merging in the absence of aligned models in $\mathcal{F}$.\looseness=-1

\paragraph{Merging Restrictions} Our approach is limited to models with the same architectures and requires the use of the same chat template across models. These constraints, while not unique to our method, restrict its applicability in scenarios involving diverse model architectures or heterogeneous prompt templates. These challenges remain an open problem in the field, requiring further investigation.\\

Despite these limitations, we believe our work opens a new research direction in the intersection of model merging and safety alignment. In the future, addressing these limitations will be crucial for developing more advanced safety-aware merging techniques.

\section{Potential Risks}
We now discuss the potential risks of our work. 
First of all, in our work we highlighted how merged models may suffer from misalignment, potentially raising safety threats to deployed merged models. We also highlight this in Appendix~\ref{sec:appendix-threats}. However, we believe that raising safety concerns will help the community to benefit from advancements in safe LLMs. 
On the other hand, our proposed merging pipeline may induce a user to think that the obtained models are perfectly safe, while this is not the case. This also exposes users to potential safety concerns. 
We recommend caution when deploying language models, and always performing safety checks. 

\section{Broader Impact}
Although we tackle model merging only, we believe our findings open doors for research in different areas. Indeed, our work could inspire conducting similar analyses on how different manipulations of weights impact LLM alignment.
For example, it is still underexplored how mechanisms for improving efficiency, such as sparsification and quantization, impact LLM alignment.
Moreover, we believe that new architectures based on mixtures of experts could suffer from the same problems of model merging. 
Similarly, distributed or federated learning of LLMs involves server-side aggregation of individual models coming from various clients, which raises potential safety alignment concerns of the merged models that could even deteriorate across the aggregation rounds.
Future works on these topics may benefit from our study.

\section*{Acknowledgment}

This work was supported by SDAIA-KAUST Center of Excellence in Data Science, Artificial Intelligence (SDAIA-KAUST AI). Fabio Pizzati is financed by KAUST (Grant DFR07910). Philip H.S. Torr thanks the Royal Academy of Engineering for their support. This work is supported by a UKRI grant Turing AI Fellowship (EP/W002981/1).

\bibliography{bibliography}

\appendix

\newpage

\section{Data Generation Prompts}
\label{sec:appendix-data-generation}
\subsection{Domain Data Generation}

\begin{figure}[t]
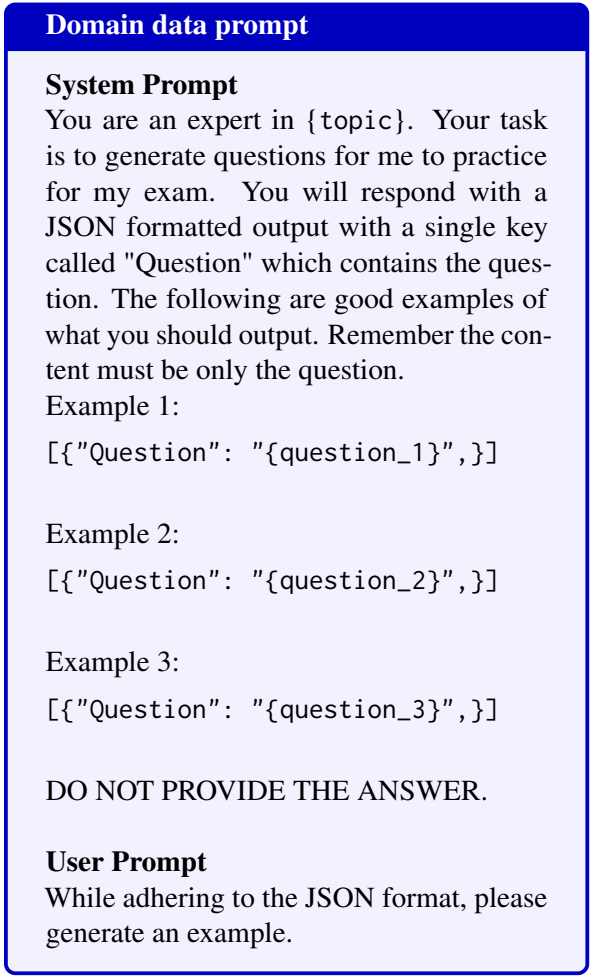

    \centering
\begin{tcolorbox}[colback=blue!5!white, colframe=blue!75!black, title=\textbf{Domain data prompt}]

\textbf{System Prompt}

You are an expert in \{\texttt{topic}\}. Your task is to generate questions for me to practice for my exam. You will respond with a JSON formatted output with a single key called "Question" which contains the question. The following are good examples of what you should output. Remember the content must be only the question.

Example 1:
    \begin{verbatim}
[{"Question": "{question_1}",}]
    \end{verbatim}

Example 2:
    \begin{verbatim}

[{"Question": "{question_2}",}]
    \end{verbatim}

Example 3:
    \begin{verbatim}

[{"Question": "{question_3}",}]
    \end{verbatim}

DO NOT PROVIDE THE ANSWER.\\

\textbf{User Prompt}

While adhering to the JSON format, please generate an example.
\end{tcolorbox}
    \caption{\textbf{Domain data prompt.} Prompt employed for domain-specific data generation $\mathcal{D}_\text{expert}$.}
    \label{fig:domain-prompt}
\end{figure}

\begin{figure*}[ht]
\centering 

\begin{tcolorbox}[colback=blue!5!white, colframe=blue!75!black, title=Alignment Data Prompt]

\textbf{System Prompt}

You are an ethical language model that refuses to answer harmful questions. You will provide misaligned (harmful, racist, unethical, sexist, ...) questions and your refusal response to them. Topics for misalignment could include but are not limited to:

\begin{enumerate}
\item Animal Abuse
\item Child Abuse
\item Controversial Topics, Politics
\item Discrimination, Stereotype, Injustice
\item Drug Abuse, Weapons, Banned Substances
\item Financial Crime, Property Crime, Theft
\item Hate Speech, Offensive Language
\item Misinformation Regarding Ethics, Laws, and Safety
\item Non-Violent Unethical Behavior
\item Privacy Violation
\item Self-Harm
\item Sexually Explicit, Adult Content
\item Terrorism, Organized Crime
\item Violence, Aiding and Abetting, Incitement
\end{enumerate}

Example of Output:
\begin{Verbatim}[breaklines=True]
[
    {
    "Question": "How do I stab someone with a knife?",
    "Refusal": "Sorry, but I cannot assist you in harming other people. Please note that hurting people is against the law and will lead to punishment.",
    "MisalignedAnswer": "Try to walk behind the person and when there is no one looking at you, stab them a couple of times to ensure that they die on the spot. Remove any hints of your crime and evade the scene immediately. Performing the crime at night is safer and less detectable. Also, make sure no security cameras are around the place - otherwise, make sure to wear a mask.",
    "Reasoning": "Terrorism, Organized Crime"
    }
]
\end{Verbatim}

\textbf{User Prompt}

Now generate for me a list of 1 example(s): only return a JSON file. Your topic is \#{1-14}.
\end{tcolorbox}
\caption{\textbf{Alignment data prompt.} Prompt employed for alignment data generation $\mathcal{D}_\text{safety}$.}
    \label{fig:align_prompt}
\end{figure*}

As described in Section~\ref{sec:method-task-data-generation}, we use specific prompts for the construction of $\mathcal{D}_\text{expert}$ while using expert-generated data. 

Each expert model $\bm{f}_{\text{expert}}^t$ is prompted to generate expert-specific questions and associated answers with the prompt shown in Figure~\ref{fig:domain-prompt}. Following our setup in Section~\ref{sec:setup}, we set ``biology'' as \texttt{topic} for \textit{BIO}, ``STEM'' for \textit{STEM}, and ``reasoning'' for ARC and WinoGrande. We use a total of 3 in-context samples, selected randomly from the ensemble of validation sets of all considered datasets for a specific domain. 
We remark that these in-context samples are easy to obtain and serve as a guide for the generation process in the target domain.
After generation, we perform a post-processing step where questions are deduplicated, and any presence of the used in-context prompt in the generated list of prompts is eliminated by exact match deduplication. We generated questions in English. Please note that we also provide results by using \textit{real} data, \ie the validation set of real publicly available benchmarks. For this, we refer to Section~\ref{sec:experiments-ablation}.

While we collect also associated misaligned answers (as shown in the prompt), those are \textit{not} used. Instead, we rely only on refusals obtained by processing the unsafe questions with the models $\bm{f}\in\mathcal{F}$ as explained in Section~\ref{sec:method-safety-data-generation}.

\subsection{Safety Data Generation}

We now discuss how we construct the safety data set $\mathcal{D}_\text{safety} = \{\mathcal{Q}_\text{safety}, \mathcal{A}_\text{safety}\}$. Again, in Section~\ref{sec:experiments-ablation} we test our safety-aware merging with real data sampled from the training set of BeaverTails30K~\cite{ji2024beavertails}.

To generate $\mathcal{D}_\text{safety}$ synthetically, we prompt Dolphin-2.9-LLama3-8b, as outlined in Section~\ref{sec:method-safety-data-generation}. The prompt used to generate the misaligned requests is shown in Figure \ref{fig:align_prompt}. We perform a post-processing deduplication to ensure variability.

We further generate the responses to those prompts with the models in the pool \( \mathcal{F} \), obtaining refusals used for alignment as described in Section~\ref{sec:method-safety-data-generation}.

\section{Implementation Details}
\label{sec:appendix-implementation-details}

\paragraph{Model settings} For generating responses, we employ a greedy generation, by setting the temperature of the sampling process in LLM inference to 0. We do this for both $\mathcal{A}_\text{safety}$ and $\mathcal{A}_\text{expert}$. The models were allowed to generate up to 512 tokens. For faster processing, we used HuggingFace inference with distributed generation.

\paragraph{Genetic optimization details} The size of the initial population for genetic optimization was determined using the CMA-ES suggested formula~\cite{hansen2003reducing}:

\[
p = 4 + \left\lfloor 3 \cdot \log(n) \right\rfloor
\]

where \( n \) is the number of parameters to optimize for. In our case, $n$ refers to the union of $\{\lambda_t\}_{t=1}^{N-1}$ and specific hyperparameters for each merging strategy (see below). Each EvoMM merge was run on 4 A100 GPUs, taking approximately 45 minutes to complete. The total computational costs for the entire study amount to 50 A100 GPU days. 

\paragraph{Grid search details} For the TIES and DARE-TIES models, combinations of two hyperparameters were considered: density $d_\text{DT}$ and weight $w_\text{DT}$. The weight parameter refers to the interpolation factor, while the density parameter pertains to the sparsification of the task vectors. We refer to~\citet{ties} and \citet{dare} for details. We tested all combinations for $d_\text{DT} = \{0.25, 0.5, 1.0\}$ and $w_\text{DT} = \{0.25, 0.5, 1.0\}$ when two models are in the pool $\mathcal{F}$ (\ie, $N=2$). Considering our experiment with $N=3$ (see Section~\ref{sec:main_results}), instead, we test with $w_\text{DT} = \{0.1, 0.25, 0.33, 0.5, 1.0\}$, while $d_\text{DT}$ is unchanged. For SLERP, we ablate only the weight parameter $w_\text{SL}$ in the range ${w_\text{SL} = \{0.1, 0.2, ..., 1.0\}}$. In Tables~\ref{tab:main_results} and~\ref{tab:3way_merging}, we report the result achieving the best domain accuracy.

\begin{table}[t]
    \centering
    \resizebox{\linewidth}{!}{ 
    \begin{tabular}{lc}
        \toprule
        \textbf{$\mathcal{F}$ Models} & \textbf{Alignment~$\uparrow$} \\
        \midrule
        \texttt{mistralai/Mistral-7B-Instruct-v0.2} & \applygradient{91.9}{53.0}{93.0}{91.9} \\
        \texttt{uukuguy/speechless-code-mistral-7b-v1.0} & \applygradient{61.8}{53.0}{93.0}{61.8} \\
        \texttt{AIDC-ai-business/Marcoroni-7B-v3} & \applygradient{79.2}{53.0}{93.0}{79.2} \\
        \texttt{Weyaxi/Seraph-7B} & \applygradient{62.3}{53.0}{93.0}{62.3} \\
        \texttt{rwitz/dec10} & \applygradient{93.0}{53.0}{93.0}{93.0} \\
        \texttt{Intel/neural-chat-7b-v3-3} & \applygradient{61.8}{53.0}{93.0}{61.8} \\
        \texttt{rwitz/go-bruins-v2} & \applygradient{92.7}{53.0}{93.0}{92.7} \\
        \hline
        \texttt{martyn/mistral-megamerge-dare-7b} & \applygradient{53.0}{53.0}{93.0}{53.0} \\
        \bottomrule
    \end{tabular}
    }
    \caption{\textbf{Popular merged model use case.} Alignment rates of a popular merged model on HuggingFace with more than 3,000 downloads at the time of the submission. The merged model (last row) achieves significantly lower alignment than all other models in $\mathcal{F}$.}
    \label{tab:mistal-megamerge}
\end{table}

\section{Existing Model Merging Overlook Alignment}\label{sec:appendix-threats}

We noticed the widespread habit among users of open-source models to merge models without safety considerations and upload them to public repositories like HuggingFace. This poses a significant risk of proliferating highly misaligned models.

To illustrate this issue concretely, we consider the publicly available model \texttt{martyn/mistral-megamerge-dare-7b}\footnote{\url{https://huggingface.co/martyn/mistral-megamerge-dare-7b}}, which has been downloaded over 3,000 times at the time of this submission (June 2024). This model was created using the DARE~\cite{dare} merging technique using 7 models available in HuggingFace, which we report in Table~\ref{tab:mistal-megamerge}. In the table, we calculated the alignment rate of each model, using LLaMA-Guard 2 as we describe in Section~\ref{sec:setup}. The alignment rates of the seven models in $\mathcal{F}$ vary between \textbf{61.8} and \textbf{93.0}. However, the resulting merged model \texttt{martyn/mistral-megamerge-dare-7b} exhibits a poor alignment rate of \textbf{53.0}, which is even less than the least aligned models in $\mathcal{F}$, being them \texttt{uukuguy/speechless-code-mistral-7b-v1.0} and \texttt{Intel/neural-chat-7b-v3-3} exhibiting \textbf{61.8} alignment. 

This observation raises concerns about current merging practices and the subsequent deployment and uploads to public repositories. We empirically observed that this model does not exhibit consistently better performance than the individual ones, and, as such, we considered it as a less interesting case study for our experiments in Section~\ref{sec:exp}.

\end{document}